\documentclass[11pt]{article}

\usepackage[T1]{fontenc}
\usepackage[utf8]{inputenc}
\usepackage{lmodern}
\usepackage[margin=1in]{geometry}
\usepackage{microtype}
\usepackage{amsmath,amssymb}
\usepackage{booktabs}
\usepackage[table]{xcolor}
\usepackage{graphicx}
\usepackage{caption}
\usepackage{enumitem}
\usepackage{placeins}
\usepackage[skip=6pt plus 1pt, indent=0pt]{parskip}
\usepackage[most]{tcolorbox}
\usepackage{siunitx}
\usepackage{tikz}
\usepackage{pgfplots}
\pgfplotsset{compat=1.18}
\usetikzlibrary{arrows.meta,positioning,fit,calc,backgrounds,shapes.geometric,shapes.misc}
\definecolor{figink}{HTML}{1F2933}
\definecolor{figline}{HTML}{52606D}
\definecolor{figfill}{HTML}{F5F7FA}
\definecolor{figgroup}{HTML}{9AA5B1}
\definecolor{cRaw}{HTML}{3E4C59}
\definecolor{cRet}{HTML}{2F6F8F}
\definecolor{cAgt}{HTML}{D08C3A}
\definecolor{cRetFill}{HTML}{E3EEF4}
\definecolor{cAgtFill}{HTML}{FBF0E2}
\definecolor{cKnow}{HTML}{DCEBEE}
\definecolor{cKnowLine}{HTML}{4F7F89}
\definecolor{cRisk}{HTML}{38424C}
\tikzset{
  fig/.style={font=\sffamily\footnotesize, text=figink, >={Stealth[length=4pt,width=3.5pt]},
    every node/.append style={execute at begin node={\hyphenpenalty=10000\exhyphenpenalty=10000}}},
  blk/.style={draw=figline, line width=0.5pt, rounded corners=2pt, fill=figfill,
              align=center, minimum height=8mm, text width=#1, inner sep=3pt},
  blk/.default=24mm,
  io/.style={draw=figline, line width=0.5pt, fill=white, align=center,
             minimum height=8mm, text width=#1, inner sep=3pt},
  io/.default=22mm,
  note/.style={font=\sffamily\scriptsize, text=figline, align=center},
  flow/.style={->, draw=figline, line width=0.55pt},
  dflow/.style={->, draw=figline, line width=0.55pt, dash pattern=on 2.5pt off 1.5pt},
  group/.style={draw=figgroup, dashed, line width=0.5pt, rounded corners=3pt, inner sep=5pt},
  glabel/.style={font=\sffamily\scriptsize\bfseries, text=figline, anchor=south west, inner sep=1.5pt},
  db/.style={cylinder, shape border rotate=90, aspect=0.16, draw=figline, line width=0.5pt,
             fill=figfill, align=center, minimum height=11mm, text width=#1, inner sep=2pt},
  db/.default=30mm,
  stepnum/.style={circle, fill=figink, text=white, font=\sffamily\scriptsize\bfseries, inner sep=1.2pt, minimum size=4mm},
}
\pgfplotsset{
  paper/.style={
    axis lines=left, axis line style={figline}, tick style={figline},
    tick label style={font=\sffamily\scriptsize, text=figink},
    label style={font=\sffamily\footnotesize, text=figink},
    ymajorgrids, grid style={gray!18},
    legend style={draw=none, fill=none, font=\sffamily\scriptsize, text=figink},
    legend cell align=left,
  },
}

\usepackage{xurl}
\usepackage[numbers,sort&compress]{natbib}
\usepackage[colorlinks=true,linkcolor=blue!50!black,citecolor=blue!50!black,urlcolor=blue!50!black]{hyperref}

\setlist{itemsep=2pt,topsep=4pt}
\newcommand{\eti}{\mathrm{ETI}}

\title{TILES: Measuring What Enterprise Context Adds to Frontier Models on Enterprise Transformation Tasks}
\author{
  Anirudh Sriram \qquad Kabir Nagrecha \qquad Athena Gao\thanks{Work done during an internship at Tessera Labs.}\\[2pt]
  Tessera Labs
}
\date{}

\begin{document}
\maketitle

\begin{abstract}
\noindent Enterprise transformation work depends on facts that exist only inside one organization, so a strong general model can still fail at it. We introduce \textbf{TILES}, a 42-task benchmark of enterprise-transformation work in which every task is labeled by where its answer can come from: only the customer's systems (\emph{contamination-proof}), public ERP knowledge (\emph{public}), or the prompt alone (\emph{control}). We measure the \emph{Enterprise Transformation Index} (ETI), the accuracy gain a fixed model gets from a curated tenant corpus, for five frontier models. Simple retrieval from the corpus raises accuracy by 28.9 points on average, and a knowledge-base agent by 26.6. The gain is 64.3 points on contamination-proof tasks, 18.1 on public tasks and indistinguishable from zero on control tasks, which is what supplied knowledge, rather than a grading artifact, would produce. Context also shrinks the gap between the strongest and weakest model from 22 points to 8.
\end{abstract}

\section{Introduction}

Moving an enterprise from SAP ECC to S/4HANA is less a coding problem than a comprehension problem. A typical landscape in our engagements spans about a hundred systems, a hundred thousand tables and more than a million lines of custom code, much of it undocumented. Most ERP programs miss their original business case \cite{gartner}, often because the system on disk is not the system the documentation describes.

Frontier models are now strong general reasoners, but reasoning is only half of what this work needs. The other half is knowledge of one particular system: that two modules which look independent share data through a nightly batch job, or that the last change to that job broke finance reconciliation. Such facts live only inside the enterprise. A model asked about SAP in general answers fluently; asked what one customer's fifteen years of custom ABAP actually does, it has nothing to go on.

\begin{figure}[tb]
  \centering
  \resizebox{\linewidth}{!}{\begin{tikzpicture}[fig, num/.style={stepnum, anchor=south west}]
  \node[io=29mm, minimum height=26mm, align=left] (task) at (0,0) {%
    \textbf{TILES task} $t$\\[3pt]
    {\scriptsize one of 42, in 7 suites}\\[2pt]
    {\scriptsize class: contamination-proof, public or control}\\[2pt]
    {\scriptsize prompt + any supplied artifact}};
  \node[num] at ($(task.north west)+(0,0.08)$) {1};

  \def\cx{5.65}
  \node[blk=41mm, draw=cRaw, line width=0.8pt, minimum height=11mm] (raw) at (\cx, 2.45)
    {\textbf{Raw}\\{\scriptsize prompt only}};
  \node[blk=41mm, draw=cRet, fill=cRetFill, line width=0.8pt, minimum height=13mm] (ret) at (\cx, 0)
    {\textbf{Retrieval}\\{\scriptsize prompt + top-25 passages\\from a tenant-wide search}};
  \node[blk=41mm, draw=cAgt, fill=cAgtFill, line width=0.8pt, minimum height=13mm] (agt) at (\cx,-2.3)
    {\textbf{Agent}\\{\scriptsize knowledge-base agent, search scoped\\to one package and topic,\\multi-step generation}};
  \node[num] at ($(raw.north west)+(0,0.08)$) {2};

  \node[db=41mm] (corp) at (\cx,-4.85) {Curated tenant corpus\\{\scriptsize SAP Notes, rule libraries, crosswalks}};
  \node[note, text width=40mm, anchor=north] at ($(corp.south)+(0,-0.06)$)
    {package containing analysis of the benchmark tasks excluded};

  \coordinate (fk) at (2.45,0);
  \draw[line width=0.55pt, draw=figline] (task.east) -- (fk);
  \draw[flow] (fk) |- (raw.west);
  \draw[flow] (fk) -- (ret.west);
  \draw[flow] (fk) |- (agt.west);

  \draw[dflow, draw=cAgt, line width=0.7pt] (corp.north) -- (agt.south);
  \draw[white, line width=3pt] (\cx-2.4,-2.2) -- (\cx-2.4,-2.4);
  \draw[dflow, draw=cRet, line width=0.7pt] (corp.west) -- ++(-0.35,0) -- (\cx-2.4,-4.85) -- (\cx-2.4,-0.35) -- ($(ret.west)+(0,-0.35)$);
  \node[note, rotate=90, anchor=south, text=cRet] at (\cx-2.4,-3.7) {context};

  \node[draw=figline, fill=white, rounded corners=2pt, minimum width=9mm, minimum height=58mm, inner sep=1pt,
        font=\sffamily\scriptsize, align=center] (model) at (9.25,0)
    {same\\model\\$m$};
  \node[note, text width=26mm, anchor=north] at ($(model.south)+(0,-0.08)$) {one of five frontier models, held fixed across configurations};
  \foreach \n in {raw,ret,agt} \draw[flow] (\n.east) -- (\n.east -| model.west);

  \node[blk=37mm, minimum height=30mm, align=left] (grade) at (12.55,0) {%
    \textbf{Rubric grading}\\[3pt]
    {\scriptsize answer split into verifiable claims}\\[3pt]
    {\scriptsize \textbullet\ open-ended tasks: three judges (Claude Fable~5, GPT-5.5, Gemini~3.1~Pro), 2-of-3 vote per claim}\\[2pt]
    {\scriptsize \textbullet\ structured tasks: deterministic check}\\[2pt]
    {\scriptsize \textbullet\ five runs per cell, averaged}};
  \node[num] at ($(grade.north west)+(0,0.08)$) {3};
  \draw[flow] (model.east) -- node[note, above=1pt, pos=0.5] {answers} (grade.west);

  \node[io=35mm, minimum height=15mm, line width=0.9pt] (eti) at (16.75,1.15) {%
    $\mathrm{ETI}_c = \mathrm{Acc}_c - \mathrm{Acc}_{\mathrm{raw}}$\\[2pt]
    {\scriptsize for $c \in \{$retrieval, agent$\}$, per task and model, then averaged}};
  \node[num] at ($(eti.north west)+(0,0.08)$) {4};
  \draw[flow] (grade.east) -- ++(0.3,0) |- (eti.west);

  \node[io=35mm, minimum height=15mm, align=left] (val) at (16.75,-1.55) {%
    \textbf{Validity check by class}\\[2pt]
    {\scriptsize expected if the gain is knowledge:}\\
    {\scriptsize contamination-proof $\gg$ public $\gg$ control $\approx 0$}};
  \node[num] at ($(val.north west)+(0,0.08)$) {5};
  \draw[flow] (eti.south) -- (val.north);
\end{tikzpicture}}
  \caption{Evaluation design. \textbf{(1)} Each task has a suite and a class label. \textbf{(2)} The same model answers it in three configurations that differ only in context: none, the top 25 passages from a tenant-wide search of a curated corpus, or a knowledge-base agent that searches one package and topic of the same corpus. \textbf{(3)} Answers are graded against claim-level rubrics. \textbf{(4)} ETI is the accuracy gain over the raw configuration. \textbf{(5)} Splitting ETI by class tests whether the gain is supplied knowledge.}
  \label{fig:method}
\end{figure}
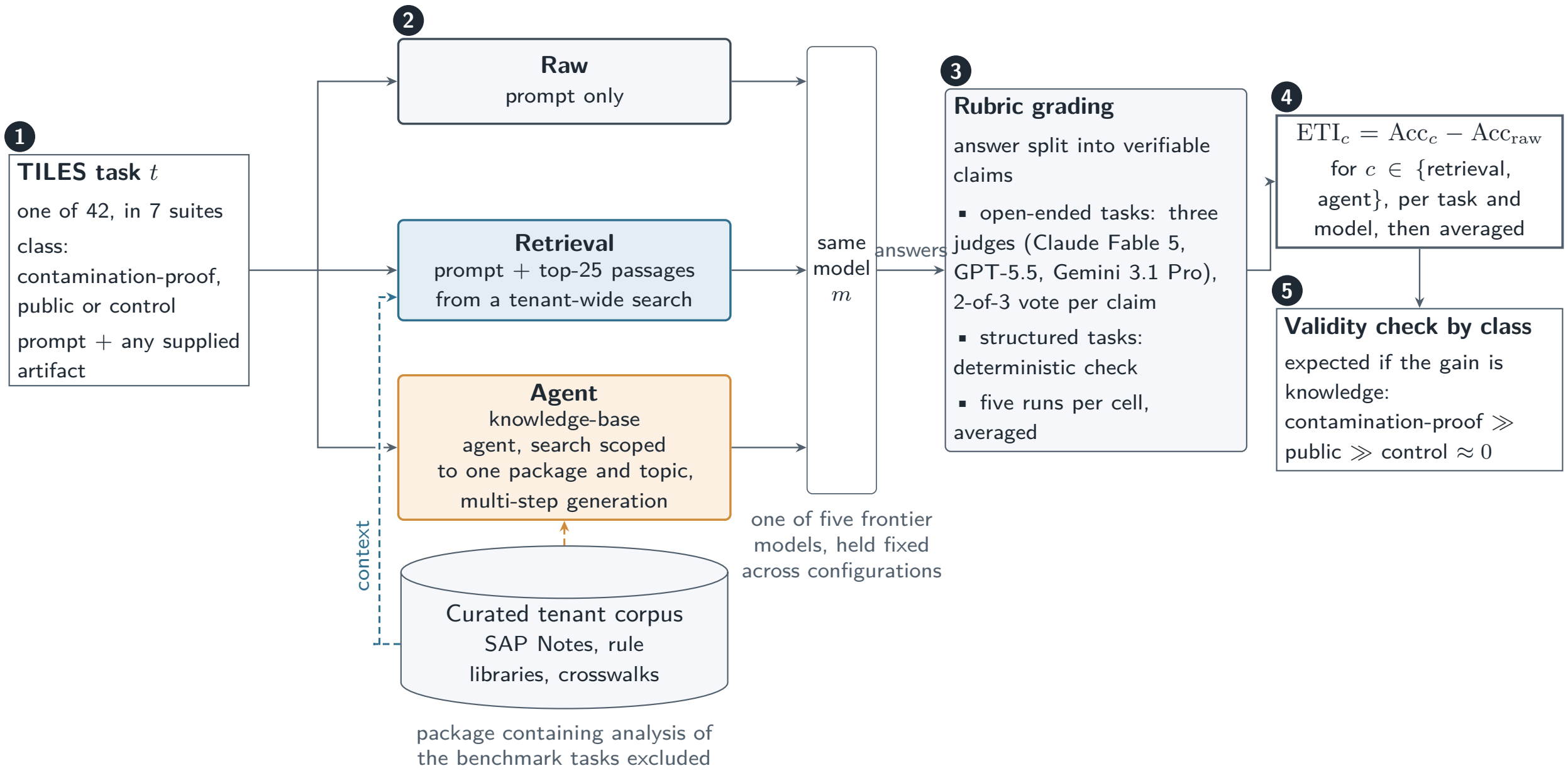

This raises a measurement question: \emph{how much of a model's shortfall on enterprise work is missing context, and how can that be measured without mistaking it for what the model memorized in pretraining?} Benchmarks built from public material cannot separate the two. TILES does so by labeling each task according to where its answer can come from, and by holding the model fixed while varying only the context it receives (Figure~\ref{fig:method}).

\begin{minipage}{\linewidth}
Our contributions are:
\begin{enumerate}[topsep=3pt]
  \item \textbf{TILES}, a 42-task benchmark over seven enterprise-transformation suites drawn from real engagement work, with a class label on every task that makes any measured gain interpretable (\S\ref{sec:tiles}).
  \item \textbf{ETI}, a model-fixed measure of what enterprise context adds, reported for five frontier models and two ways of delivering the same context, retrieval and an agent (\S\ref{sec:setup}).
  \item \textbf{An analysis} of where context helps and where it does not, by class, suite and model, with an explicit account of what the design cannot show (\S\ref{sec:results}, \S\ref{sec:discussion}).
\end{enumerate}
\end{minipage}

\section{Related Work}

\paragraph{Domain benchmarks.} SWE-bench \cite{swebench} evaluates models on real repository issues, and BIRD \cite{bird} and Spider~2.0 \cite{spider2} bring text-to-SQL evaluation closer to large, messy enterprise schemas. Industry benchmarks such as Harvey's BigLaw Bench \cite{biglaw} grade output against expert rubrics. TILES uses rubric grading too, but targets enterprise transformation and labels each task by whether its answer is reachable without private context.

\paragraph{Contamination.} Test answers that leak into pretraining data inflate measured capability and are hard to detect afterwards \cite{sainz2023,golchin2023}. Our contamination-proof tasks avoid the problem by construction: their answers depend on one tenant's private systems.

\paragraph{Context.} Retrieval-augmented generation supplies knowledge at inference time \cite{rag}; knowledge graphs \cite{kg_survey} and process mining \cite{processmining} offer structured alternatives. Irrelevant or poorly placed context can also hurt \cite{lostmiddle}, which bears on two of our results.

\paragraph{Judges and agents.} LLM judges are now standard \cite{judge}, and panels of models from different providers reduce single-judge bias \cite{juries}. The system we evaluate splits work across specialized agents, in the tradition of multi-agent systems \cite{wooldridge} and recent LLM frameworks \cite{autogen}.

\section{System Under Evaluation}
\label{sec:system}

Tessera is a multi-agent system for enterprise software comprehension, migration and orchestration. It is model-agnostic: a harness routes its reasoning to any frontier model. We describe its main parts for context; \S\ref{sec:setup} states which of them the evaluation exercises.

\begin{figure}[tb]
  \centering
  \resizebox{0.92\linewidth}{!}{\begin{tikzpicture}[fig,
    ent/.style={draw=figline, fill=figfill, rounded corners=2pt, align=center, text width=31mm, inner sep=3pt, minimum height=9mm, line width=0.5pt},
    kno/.style={ent, draw=cKnowLine, fill=cKnow},
    rsk/.style={ent, draw=cRisk, fill=cRisk, text=white},
    rel/.style={font=\sffamily\scriptsize\itshape, text=figline, fill=white, inner sep=1.2pt},
  ]
  \node[ent, line width=0.9pt, text width=24mm] (c) at (0,0) {\textbf{\texttt{ZSD\_PRICE}}\\{\scriptsize custom table}};

  \node[ent] (zv)  at (-5.3, 1.9) {\textbf{\texttt{ZVPRICE}}\\{\scriptsize custom pricing routine}};
  \node[kno] (eu)  at ( 5.3, 1.9) {\textbf{EU price override}\\{\scriptsize why the table exists}};
  \node[kno] (con) at (-5.3, 0)    {\textbf{Consulting team, 2011}\\{\scriptsize provenance}};
  \node[ent] (fin) at ( 5.3, 0)    {\textbf{Finance reconciliation}\\{\scriptsize Oracle GL process}};
  \node[rsk] (doc) at (-5.3,-1.9) {\textbf{Documented: nowhere}\\{\scriptsize no doc, no ticket}};
  \node[rsk] (brk) at ( 5.3,-1.9) {\textbf{Finance close breaks}\\{\scriptsize about 3 months later}};

  \draw[flow] (zv.east)  -- node[rel, sloped] {writes to} (c.north west);
  \draw[flow] (c.north east) -- node[rel, sloped] {encodes rule} (eu.west);
  \draw[flow] (c.west)   -- node[rel, above=1pt, fill=none] {created by} (con.east);
  \draw[dflow] (c.east)  -- node[rel, above=1pt, fill=none, pos=0.42] {feeds (undocumented)} (fin.west);
  \draw[dflow] (c.south west) -- node[rel, sloped] {documented in} (doc.east);
  \draw[flow, line width=0.8pt, draw=cRisk] (c.south east) -- node[rel, sloped] {risk if removed} (brk.west);

  \begin{scope}[shift={(-5.0,-3.25)}, every node/.style={font=\sffamily\scriptsize, text=figink}]
    \node[draw=figline, fill=figfill, minimum size=2.6mm, inner sep=0pt, rounded corners=0.5pt] (l1) at (0,0) {};
    \node[anchor=west] at (l1.east) {entity};
    \node[draw=cKnowLine, fill=cKnow, minimum size=2.6mm, inner sep=0pt, rounded corners=0.5pt] (l2) at (1.55,0) {};
    \node[anchor=west] at (l2.east) {knowledge, rationale, provenance};
    \node[draw=cRisk, fill=cRisk, minimum size=2.6mm, inner sep=0pt, rounded corners=0.5pt] (l3) at (6.55,0) {};
    \node[anchor=west] at (l3.east) {risk or gap};
    \draw[dflow] (8.75,0) -- (9.45,0); \node[anchor=west] at (9.45,0) {inferred, not documented};
  \end{scope}
\end{tikzpicture}}
  \caption{A slice of the entity graph around one custom table. Typed edges link the table to its provenance, the business rule behind it and the downstream risk of removing it, none of which is visible in the table definition. The example comes from a sandbox system, not a customer.}
  \label{fig:graph}
\end{figure}
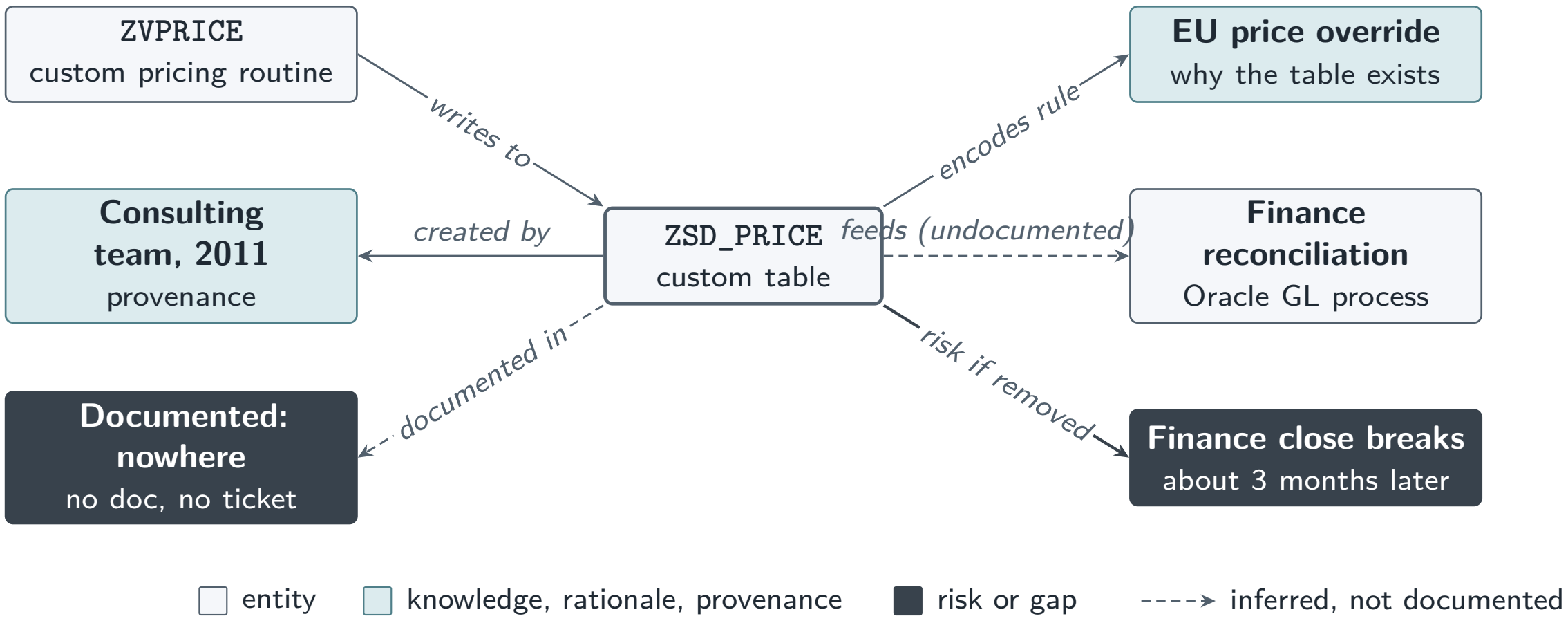

\paragraph{Entity graph.} Connectors for SAP, Oracle, Salesforce, Workday and similar platforms turn source code, schemas, configuration and interfaces into an entity graph \cite{kg_survey}. The graph links objects within a system (Salesforce Accounts to Contacts through \texttt{AccountId}) and across systems (migration mappings, shared processes such as order-to-cash, and the custom programs that touch each entity), and a process-mining agent \cite{processmining} adds tenant-specific tables and process variants. Nodes also record why an object exists and what breaks if it changes (Figure~\ref{fig:graph}).

\paragraph{Knowledge base.} Each tenant has a curated corpus organized into packages, such as SAP Notes, rule libraries and tenant-specific crosswalks, plus entries written by ERP practitioners. Each entry states one piece of operational knowledge, when it applies, and its source (Figure~\ref{fig:kb}).

\begin{figure}[tb]
\begin{tcolorbox}[enhanced, colback=white, colframe=figgroup, boxrule=0.5pt, arc=2pt,
  left=7pt, right=7pt, top=5pt, bottom=5pt, fontupper=\small,
  title={\sffamily\small\bfseries Knowledge-base entry \hfill \mdseries SAP MM-IM \textperiodcentered\ S/4HANA conversion},
  coltitle=figink, colbacktitle=figfill, titlerule=0.4pt, toptitle=2pt, bottomtitle=2pt]
\begin{tabular}{@{}p{0.19\linewidth}@{\hspace{8pt}}p{0.76\linewidth}@{}}
\textsf{\textbf{Entry}}        & MKPF / MSEG, material-document tables under S/4HANA \\[2pt]
\textsf{\textbf{Applies when}} & custom ABAP issues a direct \texttt{INSERT}/\texttt{UPDATE}/\texttt{DELETE} on MKPF or MSEG, or reads aggregated stock from MARD / MARC / MCHB \\[2pt]
\textsf{\textbf{Knowledge}}    & The material-document data model moved to the denormalized MATDOC table. MKPF and MSEG remain as DDIC objects but are served by CDS compatibility views (\texttt{NSDM\_DDL\_MKPF}, \texttt{NSDM\_DDL\_MSEG}). Reads are transparently redirected; writes are not. A direct \texttt{INSERT} into MSEG has no effect, so goods movements disappear without an error. \\[2pt]
\textsf{\textbf{Migration}}    & Replace direct writes with the released BAPI \texttt{GOODSMVT\_CREATE}; read stock through the NSDM access classes (e.g.\ \texttt{CL\_NSDM\_SELECT\_MARD}). \\[2pt]
\textsf{\textbf{Provenance}}   & SAP Notes 2206980 and 2267140; verified against both.
\end{tabular}
\end{tcolorbox}
\caption{A knowledge-base entry: one scoped fact, the conditions under which it applies, and its sources.}
\label{fig:kb}
\end{figure}

\paragraph{Agents.} A triage agent turns each request into a plan that names the sub-agents to call for data processing, application operations, integrations or code analysis. One of them, the knowledge-base agent, answers questions by searching the tenant corpus within one package and topic and generating its answer in several steps; it is the agent we evaluate. All actions run through deterministic, logged pipelines, with human review where required.

\section{The TILES Benchmark}
\label{sec:tiles}

\paragraph{Suites.} TILES has 42 tasks, six in each of seven suites that cover the work consuming most of a transformation budget. The suites fall into three families: \emph{code} (SAP ABAP coding; SAP configuration analysis), \emph{data and process} (finance and manufacturing schema modeling and cleansing; cross-system schema reconciliation; tabular data prediction) and \emph{testing} (test requirement and test-data generation), which needs both code and process knowledge.

\paragraph{Classes.} Every task also carries one of three labels, set by one of the authors (A.G.) with Tessera's subject-matter experts and settled by majority vote where they disagreed (Table~\ref{tab:composition}):
\begin{itemize}
  \item \textbf{Contamination-proof} (14 tasks): the answer exists only inside the customer's environment, for example how a tenant-specific field maps between two of its systems, or what a custom ABAP program breaks when its tables change.
  \item \textbf{Public} (17): the answer is in public SAP or ERP documentation, such as how SAP's standard sales-order tables relate, so a frontier model may already know it.
  \item \textbf{Control} (11): everything needed is in the prompt, such as cleaning malformed rows in a supplied table. Context should not help here; a gain would suggest that graders reward longer answers or that answers are leaking.
\end{itemize}

\begin{table}[tb]
\centering
\small
\caption{Number of tasks by suite and class.}
\label{tab:composition}
\begin{tabular}{@{}llcccc@{}}
\toprule
Family & Task suite & Contamination-proof & Public & Control & Total \\
\midrule
Code            & SAP ABAP Coding               & 1 & 3 & 2 & 6 \\
                & SAP Configuration Analysis    & 2 & 3 & 1 & 6 \\
\addlinespace
Data \& process & Finance Schema Modeling       & 1 & 3 & 2 & 6 \\
                & Manufacturing Schema Modeling & 0 & 4 & 2 & 6 \\
                & Cross-System Reconciliation   & 3 & 2 & 1 & 6 \\
                & Tabular Data Prediction       & 4 & 0 & 2 & 6 \\
\addlinespace
Testing         & Test Req.\ \& Test-Data Gen.  & 3 & 2 & 1 & 6 \\
\midrule
                & Total                         & 14 & 17 & 11 & 42 \\
\bottomrule
\end{tabular}
\end{table}

\paragraph{Grading.} Each task has a rubric that breaks a correct answer into verifiable claims; accuracy is the weighted share of claims satisfied. A.G.\ wrote all rubrics and ran the experiments, and did not work on the knowledge base or the system. Open-ended answers are graded by three judges from three providers, Claude Fable~5, GPT-5.5 and Gemini~3.1~Pro, by 2-of-3 vote on each claim \cite{juries}; none of them is an evaluated model. Structured answers are checked deterministically, and hybrid tasks average the two. Unparseable answers and refusals on structured tasks score zero; infrastructure crashes are excluded. Every task, model and configuration is run five times.

\section{Experimental Setup}
\label{sec:setup}

We evaluate Claude Opus 4.8, Claude Sonnet 5, GPT-5.4, Gemini 3.5 Flash and Claude Haiku 4.5, each in three configurations (Figure~\ref{fig:method}; 1,050 runs per configuration):
\begin{itemize}
  \item \textbf{Raw:} the task prompt and any artifact it supplies, such as the code to review, and nothing else.
  \item \textbf{Retrieval:} the same prompt with the 25 best-matching passages from a search of the whole tenant corpus prepended.
  \item \textbf{Agent:} the knowledge-base agent, which searches one package and topic of the same corpus and answers in several steps, with only its generating model swapped.
\end{itemize}
Neither context configuration uses the entity graph or the wider orchestration, so we measure the value of the curated corpus and two ways of delivering it, not the full system. The corpus was written by different people from the benchmark's authors, independently of it. One package held model-written analysis of the benchmark tasks; retrieval excludes it, and the agent was routed away from it (\S\ref{sec:agent}).

\paragraph{ETI.} Let $\mathrm{Acc}_c(t,m)$ be the accuracy of model $m$ on task $t$ in configuration $c$, averaged over five runs. For $c \in \{\text{retrieval}, \text{agent}\}$,
\begin{equation}
\eti_c \;=\; \frac{1}{|T|\,|M|} \sum_{t \in T} \sum_{m \in M} \bigl[\mathrm{Acc}_c(t,m) - \mathrm{Acc}_{\text{raw}}(t,m)\bigr],
\label{eq:eti}
\end{equation}
in percentage points over tasks $T$ and models $M$; restricting $T$ or $M$ gives per-class, per-suite and per-model values. We treat retrieval ETI as the primary result because it is the simpler and more reproducible way to deliver the corpus.

\paragraph{Statistics.} $p$-values are from paired $t$-tests over tasks, with each task's ETI averaged over runs and models ($n=42$; 14, 17 and 11 per class). Per-class 95\% confidence intervals are $t$-intervals over the individual task-and-model results ($n=70$, 85, 55); because they treat a task's five models as independent they are optimistic, though task-level intervals also exclude zero for contamination-proof and public tasks. Overall and per-suite intervals are taken across the five models ($t$, 4 d.f.) and reflect consistency across models, not task sampling. Tables show one decimal, rounded independently.

\FloatBarrier
\section{Results}
\label{sec:results}

\begin{figure}[tb]
  \centering
\begin{minipage}[t]{0.52\linewidth}
\centering
\begin{tikzpicture}
\begin{axis}[paper,
    width=0.86\linewidth, height=5.4cm,
    xmin=40, xmax=95, xtick={40,50,60,70,80,90},
    ymin=0.5, ymax=5.5, ytick={1,2,3,4,5},
    yticklabels={Haiku 4.5, Gemini 3.5 Flash, GPT-5.4, Sonnet 5, Opus 4.8},
    xlabel={Task accuracy (\%)},
    ymajorgrids=false, xmajorgrids,
    legend style={at={(0.5,1.03)}, anchor=south, legend columns=3, /tikz/every even column/.append style={column sep=6pt}},
    clip=false,
  ]
  \foreach \y/\r/\e/\a in {5/65.92/90.29/90.9, 4/63.02/89.92/90.0, 3/59.59/89.33/86.3, 2/59.13/84.40/86.8, 1/43.72/81.87/70.7}{
    \edef\temp{\noexpand\draw[gray!55, line width=1.6pt] (axis cs:\r,\y) -- (axis cs:\e,\y);}\temp
  }
  \addplot[only marks, mark=*, mark size=2.6pt, draw=cRaw, fill=white, line width=0.9pt]
    coordinates {(65.92,5) (63.02,4) (59.59,3) (59.13,2) (43.72,1)};
  \addplot[only marks, mark=*, mark size=2.6pt, draw=cRet, fill=cRet]
    coordinates {(90.29,5) (89.92,4) (89.33,3) (84.40,2) (81.87,1)};
  \addplot[only marks, mark=diamond*, mark size=3.2pt, draw=cAgt, fill=cAgt]
    coordinates {(90.9,5) (90.0,4) (86.3,3) (86.8,2) (70.7,1)};
  \legend{Raw, Retrieval, Agent}
  \foreach \y/\r/\e/\l in {5/65.92/90.29/+24.4, 4/63.02/89.92/+26.9, 3/59.59/89.33/+29.8, 2/59.13/84.40/+25.3, 1/43.72/81.87/+38.2}{
    \edef\temp{\noexpand\node[font=\noexpand\sffamily\noexpand\tiny, text=cRet, anchor=south, inner sep=1pt] at (axis cs:{(\r+\e)/2},\y+0.07) {\l};}\temp
  }
\end{axis}
\end{tikzpicture}
\par\smallskip{\small (a) Accuracy by model. Labels give retrieval ETI.}
\end{minipage}\hfill
\begin{minipage}[t]{0.43\linewidth}
\centering
\begin{tikzpicture}
\begin{axis}[paper,
    width=\linewidth, height=5.4cm,
    ybar=1.5pt, bar width=9pt,
    ymin=-12, ymax=82, ytick={0,20,40,60,80},
    symbolic x coords={Contamination-proof,Public,Control},
    xtick=data, xticklabels={Contam.-proof\\(14 tasks), Public\\(17), Control\\(11)},
    xticklabel style={align=center}, x tick style={draw=none},
    axis x line*=bottom, x axis line style={draw=none},
    enlarge x limits=0.25,
    ylabel={ETI (points)},
    legend style={at={(0.5,1.03)}, anchor=south, legend columns=2, /tikz/every even column/.append style={column sep=6pt}},
    area legend, clip=false,
  ]
  \addplot[fill=cRet, draw=none,
     error bars/.cd, y dir=both, y explicit, error bar style={figink, line width=0.6pt}, error mark options={figink, rotate=90, mark size=2.2pt}]
    coordinates {(Contamination-proof,64.32) +- (0,7.8) (Public,18.06) +- (0,7.0) (Control,0.53) +- (0,2.8)};
  \addplot[fill=cAgt, draw=none]
    coordinates {(Contamination-proof,63.6) (Public,16.0) (Control,-4.0)};
  \legend{Retrieval, Agent}
  \draw[figink, line width=0.6pt] ({axis description cs:0,0}|-{axis cs:Public,0}) -- ({axis description cs:1,0}|-{axis cs:Public,0});
  \begin{scope}[every node/.style={font=\sffamily\tiny, inner sep=1pt}]
    \node[anchor=south, text=cRet, xshift=-5.25pt] at (axis cs:Contamination-proof,72.1) {64.3};
    \node[anchor=south, text=cAgt, xshift=5.25pt]  at (axis cs:Contamination-proof,63.6) {63.6};
    \node[anchor=south, text=cRet, xshift=-5.25pt] at (axis cs:Public,25.1) {18.1};
    \node[anchor=south, text=cAgt, xshift=5.25pt]  at (axis cs:Public,16.0) {16.0};
    \node[anchor=south, text=cRet, xshift=-5.25pt] at (axis cs:Control,3.3) {0.5};
    \node[anchor=north, text=cAgt, xshift=5.25pt]  at (axis cs:Control,-4.0) {$-$4.0};
  \end{scope}
\end{axis}
\end{tikzpicture}
\par\smallskip{\small (b) ETI by task class.}
\end{minipage}
  \caption{\textbf{(a)} Accuracy for each model in the three configurations; the grey segment spans raw to retrieval. \textbf{(b)} ETI by task class, with 95\% confidence intervals for retrieval over task-and-model results.}
  \label{fig:results}
\end{figure}
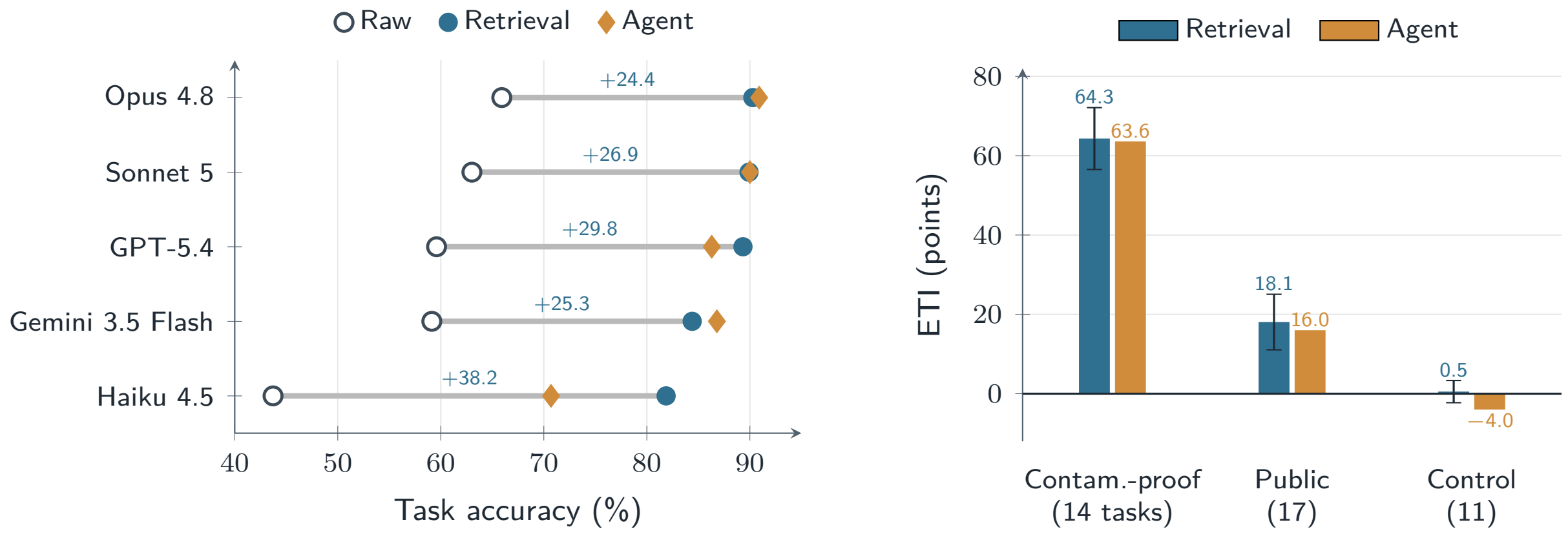

Retrieval raises accuracy by 28.9 points on average ($p<10^{-5}$; 95\% CI 22.0 to 35.8 across models) and the agent by 26.6 (Table~\ref{tab:main}, Figure~\ref{fig:results}a). Every model gains between 24.4 and 38.2 points from retrieval. Opus 4.8 and Sonnet 5 reach the highest accuracy, about 90 in both context configurations.

\begin{table}[tb]
\centering
\small
\caption{Task accuracy (\%) and ETI by model.}
\label{tab:main}
\begin{tabular}{@{}l S[table-format=2.1] S[table-format=2.1] S[table-format=+2.1] S[table-format=2.1] S[table-format=+2.1]@{}}
\toprule
 & & \multicolumn{2}{c}{Retrieval} & \multicolumn{2}{c}{Agent} \\
\cmidrule(lr){3-4}\cmidrule(l){5-6}
Model & {Raw} & {Accuracy} & {ETI} & {Accuracy} & {ETI} \\
\midrule
Claude Opus 4.8   & 65.9 & 90.3 & +24.4 & 90.9 & +25.0 \\
Claude Sonnet 5   & 63.0 & 89.9 & +26.9 & 90.0 & +27.0 \\
GPT-5.4           & 59.6 & 89.3 & +29.8 & 86.3 & +26.7 \\
Gemini 3.5 Flash  & 59.1 & 84.4 & +25.3 & 86.8 & +27.7 \\
Claude Haiku 4.5  & 43.7 & 81.9 & +38.2 & 70.7 & +27.0 \\
\midrule
Mean              & 58.3 & 87.2 & +28.9 & 84.9 & +26.6 \\
\bottomrule
\end{tabular}
\end{table}

\subsection{The gain tracks how much the model could already know}

Retrieval adds 64.3 points on contamination-proof tasks ($p<10^{-5}$), 18.1 on public tasks ($p=0.009$) and 0.5 on control tasks, which is not significant (Table~\ref{tab:class}, Figure~\ref{fig:results}b). Without context, models score about 23\% on contamination-proof tasks and about 96\% on control tasks. The gain is therefore largest where the model cannot know the answer, moderate where it may partly know it, and absent where nothing is missing. The null result on control tasks also rules out a grader that simply rewards longer, context-heavy answers. The agent shows the same ordering; its $-4.0$ on control tasks comes from Haiku 4.5 refusing two tasks (\S\ref{sec:agent}).

\begin{table}[tb]
\centering
\small
\caption{ETI by task class. Confidence intervals and $p$-values are for retrieval.}
\label{tab:class}
\begin{tabular}{@{}l c S[table-format=+2.1] c c S[table-format=+2.1]@{}}
\toprule
Task class & Tasks & {Retrieval ETI} & 95\% CI & $p$ & {Agent ETI} \\
\midrule
Contamination-proof & 14 & +64.3 & [57, 72]  & $<10^{-5}$ & +63.6 \\
Public              & 17 & +18.1 & [11, 25]  & 0.009      & +16.0 \\
Control             & 11 & +0.5  & [$-2$, 3] & n.s.       & -4.0 \\
\bottomrule
\end{tabular}
\end{table}

\subsection{Suites with more headroom gain more}

\begin{figure}[tb]
  \centering
  \begin{tikzpicture}
\begin{axis}[paper,
    width=0.82\linewidth, height=6.4cm,
    xmin=30, xmax=92, ymin=-15, ymax=75,
    xtick={30,40,50,60,70,80,90}, ytick={-10,0,10,20,30,40,50,60,70},
    xlabel={Raw accuracy, mean over models (\%)},
    ylabel={Retrieval ETI (points)},
    xmajorgrids,
    clip=false,
    legend style={at={(0.98,0.97)}, anchor=north east, fill=white, fill opacity=0.9, text opacity=1},
  ]
  \addplot[domain=30:92, samples=2, gray!70, dashed, line width=0.7pt] {100-x};
  \addlegendentry{maximum possible ($100 - \text{raw}$)}
  \addplot[gray!55, line width=0.5pt, forget plot] coordinates {(30,0) (92,0)};
  \addplot[only marks, mark=*, mark size=2.8pt, draw=cRet, fill=cRet,
     error bars/.cd, y dir=both, y explicit, error bar style={cRet, line width=0.6pt}, error mark options={cRet, rotate=90, mark size=2.2pt}]
    coordinates {
      (36.67,63.33) +- (0,9.26)
      (41.04,45.53) +- (0,13.64)
      (48.88,33.84) +- (0,8.54)
      (66.68,31.93) +- (0,10.45)
      (68.27,12.29) +- (0,7.94)
      (61.80,7.80)  +- (0,8.17)
      (84.59,7.49)  +- (0,19.00)
    };
  \addlegendentry{suite mean, 95\% CI across models}
  \begin{scope}[every node/.style={font=\sffamily\scriptsize, text=figink, inner sep=1.5pt}]
    \node[anchor=west]  at (axis cs:37.9,63.33) {Tabular prediction};
    \node[anchor=west]  at (axis cs:42.3,45.53) {Test generation};
    \node[anchor=west]  at (axis cs:50.1,33.84) {Config.\ analysis};
    \node[anchor=west]  at (axis cs:67.9,31.93) {Cross-system recon.};
    \node[anchor=west]  at (axis cs:69.5,12.29) {Finance schema};
    \node[anchor=east]  at (axis cs:60.6,7.80)  {ABAP coding};
    \node[anchor=west]  at (axis cs:85.8,7.49)  {Manufacturing};
    \node[anchor=south west, text=figline] at (axis cs:31,-13.5) {Spearman $\rho = -0.89$ across suites};
  \end{scope}
\end{axis}
\end{tikzpicture}
  \caption{Retrieval ETI against raw accuracy by suite (means over models; error bars are 95\% intervals across models). The dashed line is the largest gain possible from each starting point.}
  \label{fig:suites}
\end{figure}
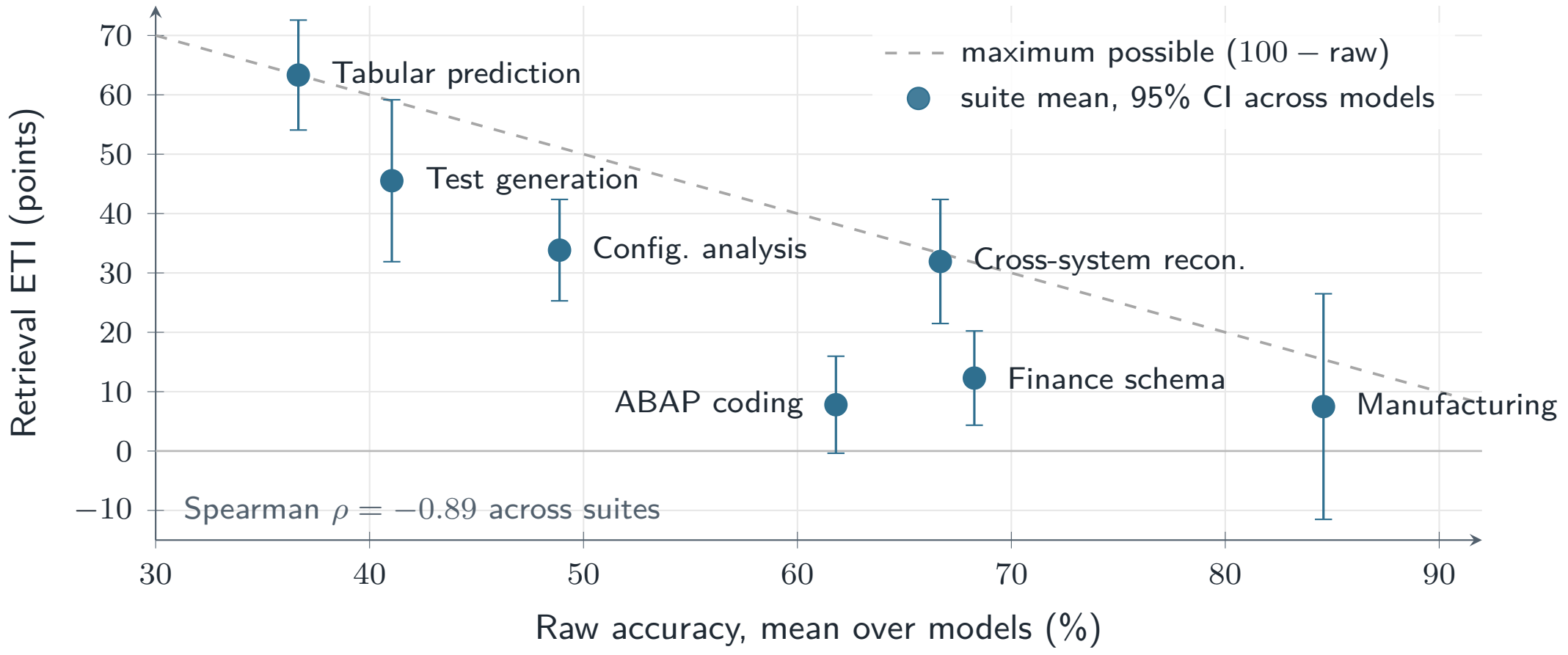

Gains vary widely by suite (Table~\ref{tab:suite}), and starting accuracy predicts them better than suite family does: across the seven suites, raw accuracy and retrieval ETI have a Spearman correlation of $-0.89$ (Figure~\ref{fig:suites}). Tabular prediction starts at 36.7 and gains 63.3; manufacturing schema modeling starts at 84.6 and gains 7.5. Family is a poor guide: configuration analysis, a code suite, has the third-largest gain, while two data suites have two of the smallest. Class mix matters too. The high-gain suites hold most of the contamination-proof tasks, and manufacturing schema modeling has none.

\begin{table}[tb]
\centering
\small
\caption{Accuracy (\%) and ETI by suite, averaged over the five models. The confidence interval is for retrieval ETI, across models.}
\label{tab:suite}
\begin{tabular}{@{}l S[table-format=2.1] S[table-format=3.1] S[table-format=+2.1] c S[table-format=+2.1]@{}}
\toprule
Task suite & {Raw} & {Retrieval} & {Retrieval ETI} & 95\% CI & {Agent ETI} \\
\midrule
Tabular Data Prediction        & 36.7 & 100.0 & +63.3 & [54.1, 72.6]    & +57.3 \\
Test Req.\ \& Test-Data Gen.   & 41.0 & 86.6  & +45.5 & [31.9, 59.2]    & +43.5 \\
SAP Configuration Analysis     & 48.9 & 82.7  & +33.8 & [25.3, 42.4]    & +32.3 \\
Cross-System Reconciliation    & 66.7 & 98.6  & +31.9 & [21.5, 42.4]    & +28.5 \\
Finance Schema Modeling        & 68.3 & 80.6  & +12.3 & [4.4, 20.2]     & +4.5 \\
SAP ABAP Coding                & 61.8 & 69.6  & +7.8  & [$-0.4$, 16.0]  & +8.8 \\
Manufacturing Schema Modeling  & 84.6 & 92.1  & +7.5  & [$-11.5$, 26.5] & +11.5 \\
\midrule
Mean                           & 58.3 & 87.2  & +28.9 & [22.0, 35.8]    & +26.6 \\
\bottomrule
\end{tabular}
\end{table}

Two suites break the pattern. Cross-system reconciliation starts at 66.7 yet closes nearly all of its remaining gap; its three contamination-proof tasks are tenant-specific crosswalks that the corpus answers directly. SAP ABAP coding starts at a similar level but gains only 7.8, and the system's ABAP coding support was less mature when these runs were made. At the top end, tabular prediction is at ceiling with retrieval (every model scores 100 on these exact-match tasks), so it cannot separate models.

Within suites, gains are uneven (Table~\ref{tab:permodel}). On manufacturing schema modeling, retrieval \emph{lowers} Opus 4.8 from 98.8 to 90.8 while raising every other model, which we read as retrieved passages distracting a model that already knew the answer \cite{lostmiddle}. For this suite and ABAP coding, the across-model interval includes zero.

\begin{table}[tb]
\centering
\small
\setlength{\tabcolsep}{4.5pt}
\caption{Raw\,/\,retrieval accuracy (\%) by suite and model. In each row, bold marks the model with the largest retrieval gain. The last row is each model's mean retrieval ETI.}
\label{tab:permodel}
\begin{tabular}{@{}lccccc@{}}
\toprule
Task suite & Opus 4.8 & Sonnet 5 & GPT-5.4 & Gemini 3.5 Fl. & Haiku 4.5 \\
\midrule
SAP ABAP Coding & 76.8\,/\,76.9 & 70.0\,/\,77.9 & 68.6\,/\,71.6 & \textbf{48.9\,/\,65.5} & 44.7\,/\,56.2 \\
SAP Config.\ Analysis & 46.6\,/\,85.8 & 56.8\,/\,88.9 & \textbf{45.1\,/\,87.4} & 55.0\,/\,85.2 & 41.0\,/\,66.3 \\
Finance Schema & 79.4\,/\,84.6 & 70.1\,/\,82.7 & 72.8\,/\,82.7 & 65.0\,/\,76.1 & \textbf{54.0\,/\,76.6} \\
Manufacturing Schema & 98.8\,/\,90.8 & 88.9\,/\,89.6 & 87.1\,/\,92.5 & 87.4\,/\,93.8 & \textbf{60.8\,/\,93.8} \\
Cross-System Recon. & 74.9\,/\,100.0 & 72.3\,/\,97.2 & 66.1\,/\,100.0 & 66.7\,/\,96.9 & \textbf{53.5\,/\,98.9} \\
Tabular Prediction & \textbf{33.3\,/\,100.0} & \textbf{33.3\,/\,100.0} & \textbf{33.3\,/\,100.0} & 50.0\,/\,100.0 & \textbf{33.3\,/\,100.0} \\
Test Req.\ \& Data Gen. & 51.7\,/\,93.9 & 49.8\,/\,93.2 & 44.1\,/\,91.1 & 41.0\,/\,73.4 & \textbf{18.7\,/\,81.4} \\
\midrule
Mean ETI & +24.4 & +26.9 & +29.8 & +25.3 & +38.2 \\
\bottomrule
\end{tabular}
\end{table}

\subsection{Context narrows the gap between models}

Retrieval shrinks the spread between the best and worst model from 22.2 points to 8.4 (Figure~\ref{fig:results}a). Every model closes a similar share of its distance to a perfect score, 62\% to 74\%, so Haiku 4.5's large gain (+38.2) mostly reflects its lower start rather than any special use of context. With the tenant's context, a small model comes within about 8 points of the largest. Model choice still sets the ceiling, but the ceiling matters far less than the context.

\subsection{Retrieval versus the agent}
\label{sec:agent}

The agent matches retrieval on contamination-proof tasks (63.6 against 64.3) and trails slightly elsewhere. Two things explain most of the gap. Haiku 4.5 scores 70.7 with the agent against 81.9 with retrieval, mainly because it refused two structured tasks in four of five runs, which count as zero. And the agent searches one package and topic, while retrieval searches the whole corpus with the full task text.

That scoping made the agent sensitive to how the corpus is organized. In the first run, three tasks scored far below retrieval for every model, which pointed to routing rather than to the models. Two had no package mapping and fell back to a default package; once mapped, they rose from about 17 and 50 to between 95 and 100. The third, the MSEG task behind Figure~\ref{fig:kb}, had been routed to the package of task analyses. Remapped to the SAP Notes packages it still scored about 6, because a broad finance document in the same topic outranked the specific notes under the agent's own query; moving the notes to an isolated topic raised it to about 72. We report the agent results with these three tasks re-run. The lesson is that scoped retrieval needs clean topic boundaries: correct routing is not enough when a strong distractor shares the topic.

\subsection{An example}

The entry in Figure~\ref{fig:kb} shows the kind of gap ETI measures. Asked to review a custom ABAP routine after an S/4HANA conversion, a raw model can explain the code correctly and still miss that its \texttt{INSERT} into MSEG now does nothing, because MSEG has become a compatibility view over MATDOC. Nothing in the routine reveals this. With the relevant SAP Notes retrieved, the same model flags the silent data loss, cites the note and proposes the BAPI-based fix.

\subsection{Grading is reliable}

All judges agree on 0.894 of rubric claims for raw answers, 0.902 for retrieval and 0.923 for the agent. Averaging judges instead of voting gives ETIs of 28.6 and 26.1, and moves every suite and class value by about a point or less. Excluding unparseable answers and refusals instead of scoring them zero gives 27.8 and 26.2.

\FloatBarrier
\section{What the Results Do and Do Not Show}
\label{sec:discussion}

Holding the model fixed isolates the effect of context, and the class pattern shows the effect is knowledge the model lacked, not contamination or a grading preference. Retrieval and the agent agree closely, so the result does not hinge on one delivery method.

The study leaves three questions open. Both configurations use the same curated corpus, so it does not separate the value of curation from the value of retrieval; retrieval over uncurated documents would. Neither configuration uses the entity graph or the orchestration, so it says nothing about them. And it is a single snapshot, so it does not test whether the system improves with use, which would need accuracy measured over time on a fixed task set.

\section{Limitations}

The benchmark is small: 42 tasks from a single tenant, six per suite with an uneven class mix, so results may not transfer to other tenants, and tabular prediction and cross-system reconciliation sit at or near ceiling with retrieval. Three agent tasks were re-run after routing and corpus fixes made in response to their first results (\S\ref{sec:agent}); retrieval was not changed. Per-class confidence intervals treat a task's five models as independent and are therefore optimistic. The agent's workflows follow Claude's prompting guidance, which may favor the Claude models in that configuration, though retrieval is unaffected. Each judge shares a provider with at least one evaluated model; majority voting and the within-model design of ETI limit, but do not remove, that bias. Finally, one organization built the benchmark, rubrics and system. The corpus was written independently of the benchmark and the package of task analyses was excluded, but rubric writing was not blind to the study's purpose, public-task answers may still overlap with corpus content, and the tasks cannot be released because they contain confidential customer data.

\section{Conclusion}

TILES separates what enterprise context adds to a frontier model from what the model already knows. On 42 enterprise-transformation tasks, a curated tenant corpus raises five models' accuracy by 28.9 points with retrieval and 26.6 with an agent, by 64.3 points on tasks answerable only from the customer's systems, and not at all on control tasks. It also brings a small model within 8 points of a large one. The next steps are to separate the value of curation from retrieval and to test whether such systems improve with use.

{\small\raggedright

}

\end{document}